\documentclass[runningheads,a4paper]{llncs}

\usepackage{caption}
\usepackage{amssymb}
\usepackage{amsmath}
\usepackage{graphicx}
\usepackage{mathtools}
\usepackage{hyperref}
\usepackage{url}
\urldef{\mailsa}\path|kratarthgoel@gmail.com, |
\urldef{\mailsb}\path|ronvohra@gmail.com,|
\urldef{\mailsc}\path|jksahoo@goa.bits-pilani.ac.in|

\newcommand{\keywords}[1]{\par\addvspace\baselineskip
\noindent\keywordname\enspace\ignorespaces#1}

\begin{document}

\mainmatter  

\title{Polyphonic Music Generation by Modeling Temporal Dependencies Using a RNN-DBN}

\titlerunning{Polyphonic Music Generation Using a RNN-DBN}

%
%
\author{Kratarth Goel%
\and Raunaq Vohra\and J. K. Sahoo}
\authorrunning{Kratarth Goel, Raunaq Vohra and J. K. Sahoo}

\institute{Birla Institute of Technology and Science, Pilani,\\
Goa, India\\
\mailsa\\
\mailsb\\
\mailsc\\}

%
%

\toctitle{Lecture Notes in Computer Science}
\tocauthor{Authors' Instructions}
\maketitle

\begin{abstract}
\emph{In this paper, we propose a generic technique to model temporal dependencies and sequences using a combination of a recurrent neural network and a Deep Belief Network. Our technique, RNN-DBN, is an amalgamation of the memory state of the RNN that allows it to provide temporal information and a multi-layer DBN that helps in high level representation of the data. This makes RNN-DBNs ideal for sequence generation. Further, the use of a DBN in conjunction with the RNN makes this model capable of significantly more complex data representation than an RBM. We apply this technique to the task of polyphonic music generation.}
\keywords{Deep architectures, recurrent neural networks, music generation, creative machine learning, Deep Belief Networks, generative models}
\end{abstract}

\section{Introduction}

Creative machine learning is an extremely relevant research area today. Generative models form the basis of creative learning, and for this reason, a high level of sophistication is required from these models.  Also, identifying features in the subjective fields of art, literature and music is an arduous task, which is only made more difficult when more elaborate learning is desired.  Deep architectures, therefore, present themselves as an ideal framework for generative models, as they are inherently stochastic and support increasingly complex representations with each added layer. Recurrent neural networks (RNNs) have also used with great success as regards generative models, particularly handwriting generation, where they have been used to achieve the current state-of-the-art results. The internal feedback or memory state of these neural networks is what makes them a suitable technique for sequence generation in tasks like polyphonic music composition. 

There have been many attempts to generate polyphonic music in the past, and a variety of techniques have been applied, some of which will be discussed here. Matic \emph{et al.} \cite{A13} used neural networks and cellular automata to generate melodies. However, this is dependent on a feature set based on emotions and further requires specialized knowledge of musical theory. A similar situation is observed with the work done by Maeda and Kajihara \cite{A17}, who used genetic algorithms for music generation. Elman networks with chaotic inspiration were used by \cite{A15} to compose music. While RNNs are an excellent technique to model sequences, they used chaotic inspiration as an external input instead of real stochasticity to compose original music. A Deep Belief Network with a sliding window mechanism to create jazz melodies was proposed by \cite{A37}. However, due to lack of temporal information, there were many instances of repeated notes and pitches. Most recently, Boulanger-Lewandowski \emph{et al.} \cite{A11} used RNN-RBMs for polyphonic music generation and obtained promising results. We propose a generic technique which is a combination of a RNN and a Deep Belief Network for sequence generation, and apply it to automatic music composition. Our technique, the \emph {RNN-DBN}, effectively combines the superior sequence modeling capabilities of a RNN with the high level data modeling that a DBN enables to produce rich, complex melodies which do not feature significant repetition. Moreover, the composition of these melodies does not require any feature selection from the input data. This model is presesnted as a generic technique, \emph{i. e.} it does not make any assumptions about the nature of the data. We apply our technique to a variety of datasets and have achieved excellent results which are on par with the current state-of-the-art. 

The rest of this paper is organized as follows: Section 2 discusses various deep and neural network architectures that serve as the motivation for our technique, described in Section 3.  We demonstrate the application of RNN-DBNs to the task of polyphonic music generation in Section 4 and present our results. Section 5 discusses possible future work regarding our technique and concludes the paper.

\section{Preliminaries}

\subsection{Restricted Boltzmann Machines}

Restricted Boltzmann machines (RBMs) are energy based models with their energy function $E(v,h)$ defined as:

\begin{equation}
E(v,h) = - b'v - c'h - h'Wv
\end{equation}

where W represents the weights connecting hidden $(h)$ and visible $(v)$ units and $b$, $c$ are the biases of the visible and hidden layers respectively.

This translates directly to the following free energy formula:

\begin{equation}
\mathcal{F}(v)= - b'v - \sum_i \log \sum_{h_i} e^{h_i (c_i + W_i v)}.
\end{equation}

Because of the specific structure of RBMs, visible and hidden units are conditionally independent given one another. Using this property, we can write:

\begin{equation}
p(h|v) = \prod_i p(h_i|v)
\end{equation}
\begin{equation}
p(v|h) = \prod_j p(v_j|h).
\end{equation}

Samples can be obtained from a RBM by performing block Gibbs sampling, where visible units are sampled simultaneously given fixed values of the hidden units. Similarly, hidden units are sampled simultaneously given the visible unit values. A single step in the Markov chain is thus taken as follows,

\begin{equation}
\begin{split}
h^{(n+1)} = \sigma(W'v^{(n)} + c) \\
v^{(n+1)} = \sigma(W h^{(n+1)} + b),
\end{split}
\end{equation}

where $\sigma$ represents the sigmoid function acting on the activations of the $(n + 1)^{th}$ hidden and visible units. Several algorithms have been devised for RBMs in order to efficiently sample from $p(v,h)$ during the learning process, the most effective being the well-known \emph{contrastive divergence} $(CD-k)$ algorithm \cite{A42}.

In the commonly studied case of using binary units $($where $v_j$ and $h_i$ $\in
\{0,1\})$, we obtain, from Eqn. (4), a probabilistic version of the activation function:

\begin{equation}
\begin{split}
P(h_i=1|v) = \sigma(c_i + W_i v) \\
P(v_j=1|h) = \sigma(b_j + W'_j h)
\end{split}
\end{equation}

The free energy of an RBM with binary units thus further simplifies to:

\begin{equation}
\mathcal{F}(v)= - b'v - \sum_i \log(1 + e^{(c_i + W_i v)}).
\end{equation}

We obtain the following log-likelihood gradients for an RBM with binary units:

\begin{equation}
\begin{multlined}
- \frac{\partial{ \log p(v)}}{\partial W_{ij}} =
    E_v[p(h_i|v) \cdot v_j]
    - v^{(i)}_j \cdot \sigma(W_i \cdot v^{(i)} + c_i) \\
-\frac{\partial{ \log p(v)}}{\partial c_i} =
    E_v[p(h_i|v)] - \sigma(W_i \cdot v^{(i)}) \shoveleft[1.4cm] \\
-\frac{\partial{ \log p(v)}}{\partial b_j} =
    E_v[p(v_j|h)] - v^{(i)}_j\shoveleft[1.5cm]
\end{multlined}
\end{equation}

where $\sigma(x) = (1 + e ^{-x})^{-1}$ is the element-wise logistic
sigmoid function. 

\subsection{Recurrent Neural Network}

\emph{Recurrent neural networks} (RNNs) are a particular family of neural networks where the network contains one or more feedback connections, so that activation of a cluster of neurons can flow in a loop. This property allows for the network to effectively model time series data and learn sequences. An interesting property of RNNs is that they can be modeled as feedforward neural networks by unfolding them over time. RNNs can be trained using the \emph{Backpropogation Through Time} (BPTT) technique. If a network training sequence starts at a time instant $t_0$ and ends at time $t_1$, the total cost function is simply the sum over the standard error function $E_{sse/ce}$ at each time step, 
\begin{equation}
	E_{total} = \sum\nolimits_{t=t_0}^{t_1} E_{sse/ce}(t)
\end{equation}
and the gradient descent weight update contributions for each time step are given by,

\begin{equation}
	\Delta w_{ij} = -\eta \frac{\partial E_{total}(t_0,t_1)}{\partial w_{ij}} =  \sum\nolimits_{t=t_0}^{t_1} \frac{\partial E_{sse/ce}(t)}{\partial w_{ij}}
\end{equation}

The partial derivatives of each component $\frac{\partial E_{sse/ce}}{\partial w_{ij}}$ now have contributions from multiple instances of each weight $w_{ij} \in \{W_{v^{(t-1)}h^{(t-1)}}, W_{(h^{t-1)}h^{(t)}}\}$ and are dependent on the inputs and hidden unit activations at previous time instants. The errors now must be back-propagated through time as well as through the network.

\subsection{Deep Belief Network}
RBMs can be stacked and trained greedily to form \emph{Deep Belief Networks} (DBNs). DBNs are graphical models which learn to extract a deep hierarchical representation of the training data \cite{A05}. They model the joint distribution between observed vector $\mathbf{x}$ and the $\ell$ hidden layers $h^k$ as follows:

\begin{equation}
P(\mathbf{x}, h^1, \ldots, h^{\ell}) = \left(\prod_{k=0}^{\ell-2} P(h^k|h^{k+1})\right) P(h^{\ell-1},h^{\ell})
\end{equation}
where $\mathbf{x}=h^0$, $P(h^{k-1} | h^k)$ is a conditional distribution for the visible units conditioned on the hidden units of the RBM at level $k$, and $P(h^{\ell-1}, h^{\ell})$ is the visible-hidden joint distribution in the top-level RBM. 

The principle of greedy layer-wise unsupervised training can be applied to DBNs with RBMs as the building blocks for each layer \cite{A35}.
We begin by training the first layer as an RBM that models the raw input $x = h^{(0)}$ as its visible layer. Using that first layer, we obtain a representation of the input that will be used as data for the second layer. Two common solutions exist here, and the representation can be chosen as the mean activations $p(h^{(1)}=1|h^{(0)})$ or samples of $p(h^{(1)}|h^{(0)})$. Then we train the second layer as an RBM, taking the transformed data (samples or mean activations) as training examples (for the visible layer of that RBM). In the same vein, we can continue adding as many hidden layers as required, while each time propagating upward either samples or mean values.

\subsection{Recurrent Temporal Restricted Boltzmann Machine}
The \emph{Recurrent Temporal Restricted Boltzmann Machine} (RTRBM) \cite{A43} is a sequence of conditional RBMs (one at each time instant) whose
parameters $\{b_v^{t} $, $b_h^{t}$ , $W^{t} \}$ are time-dependent and depend on the sequence history at time $t$, denoted by $\mathbf{A(t)}= \{v^{(\tau)} , u^{(\tau)}  | \tau < t\}$,  where $u^{(t)}$ is the mean-field value of $h(t)$, as seen in \cite{A11}. The RTRBM is formally defined by its joint probability distribution,

\begin{equation}
 P ({v^{(t)}, h(t)}) =  \prod_{t=1}^{T} {P (v^{(t)} , h^{(t)} |\bold{A^{(t)}})}
\end{equation}

where $P (v^{(t)} , h^{(t)} |\bold{A^{(t)}})$ is the joint probability
 of the $t^{th}$ RBM whose parameters are defined below, from Eqn. (13) and Eqn. (14).
 While all the parameters of the RBMs may usually depend on the previous time instants, we will consider the case where only the biases depend on $u^{(t-1)}$.
 
\begin{equation}
b_h^{(t)} = b_h + W_{uh}u^{(t-1)}
\end{equation}
\begin{equation}
b_v^{(t)} = b_v + W_{uv}u^{(t-1)}
\end{equation}

which gives the RTRBM six parameters,$\{ W, b_v , b_h , W_{uv} , W_{uh} , u^{(0)}\}$. The more general scenario is derived in similar fashion.  While the hidden units $h^{(t)}$ are binary during inference and sample generation, it is the \emph{mean-field} value $u^{(t)}$ that is transmitted to its successors (Eqn. (15)). This important distinction makes exact inference of the $u^{(t)}$ easy and improves the efficiency of training\cite{A43}.
 
\begin{equation}
u^{(t)} = \sigma(W_{vh}v^{(t)} + b_h^{(t)} ) = \sigma(W_{vh}v^{(t)} + W_{uh} u^{(t-1)} + b_h )
\end{equation}

Observe that Eqn. (15) is exactly the defining equation of a RNN (defined in Section 4) with hidden units $u^{(t)}$ .

\section{RNN-DBN}

The RTRBM can be thought of as a sequence of conditional RBMs whose parameters are the output of a deterministic RNN \cite{A11}, with the constraint that the hidden units must describe the conditional distributions and convey temporal information for sequence generation. The use of a single RBM layer greatly constricts the expressive power of the model as a whole. This constraint can be lifted by combining a full RNN having distinct hidden units $u^{(t)}$ with a RTRBM graphical model, replacing the RBM structure with the much more powerful model of a DBN. We call this model the RNN-DBN. 

In general, the parameters of the RNN-DBN are made to depend only on $u^{(t-1)}$ given by Eqn. (13) and Eqn. (14) along with,
\begin{equation}
b_{h_2}^{(t)} = b_{h_2} + W_{uh_2}u^{(t-1)}
\end{equation}

for the second hidden layer in an RNN-DBN with 2 hidden layers. The joint probability distribution of the RNN-DBN is also given by Eqn. (12), but with $u^{(t)}$ defined arbitrarily, as given by Eqn. (17). For simplicity, we consider the RNN-DBN parameters to be $\{W_{vh^{(t)}}$, $b_v^{(t)}$ , $b_{h_1}^{(t)}$ , $b^{(t)}_{h_2}\}$ for a 2 hidden layer RNN-DBN (shown in Fig. 1), i.e. only the biases are variable, and a single-layer RNN, whose hidden units $u^{(t)}$ are only connected to their direct predecessor $u^{(t-1)}$ and to $v^{(t)}$ by the relation,

\begin{equation}
u^{(t)} =  \sigma(W_{vu}v^{(t)} + W_{uu}h ^{(t -1)}+ b_{u}).
\end{equation}

\begin{figure}
\centering
\includegraphics[height=6.2cm]{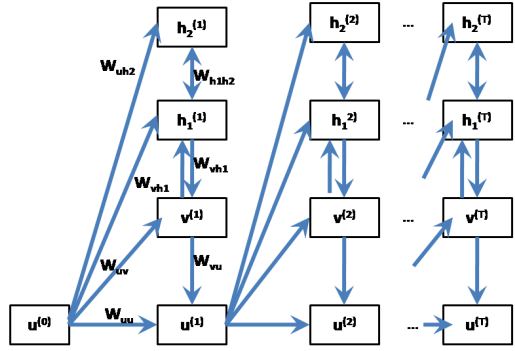}
\caption{A RNN-DBN with 2 hidden layers.}
\label{fig:example}
\end{figure}

The DBN portion of the RNN-DBN is otherwise exactly the same as any general DBN. This gives the 2 hidden layer RNN-DBN twelve parameters, $\{ $ $W_{vh_1}$,$W_{h_1h_2}$, $b_v$ , $b_{h_1}$ , $b_{h_2}$, $W_{uv}$ , $W_{uh_1}$ , $W_{uh_2}$, $u^{(0)}$, $W_{vu}$, $W_{uu}$ , $b_u$ $\}$.\\

The training algorithm is based on the following general scheme:

\begin{enumerate}
\item Propagate the current values of the hidden units $u^{(t)}$ in the RNN portion of the graph using Eqn. (17).
\item Calculate the DBN parameters that depend on the $u^{(t)}$
 (Eqn. (13), (14) and (16)) by greedily training layer-by-layer of the DBN, each layer as an RBM (Train the first layer as an RBM that models the raw input as its visible layer and use that first layer to obtain a representation of the input that will be used as data for the second layer and so on).
\item Use CD-k to estimate the log-likelihood gradient (Eq. (8)) with respect to $W$ , $b_v$ and $b_h$ for each RBM composing the DBN.
\item Repeat steps 2 and 3 for each layer of the DBN.
\item Propagate the estimated gradient with respect to
$b_v^{(t)}$ , $b_h^{(t)}$ and $b_{h_2}^{(t)}$  backward through time (BPTT) \cite{A44} to obtain the estimated gradient
with respect to the RNN, for the RNN-DBN with 2 hidden layers.

\end{enumerate}

\section{Implementation and Results}

\begin{center}

\captionof{table}{Log-likelihood (LL) for various musical models in the polyphonic music generation task.}

	\begin{tabular}{|p{3cm}|p{2.35cm}|p{1.85cm}|p{2.15cm}|p{2.45cm}|}
	

		\hline
		\bfseries{Model} & \bfseries{JSB Chorales (LL)} & \bfseries{MuseData (LL)} & \bfseries{Nottingham (LL)} & \bfseries{Piano-Midi.de (LL)}\\
		\hline
		Random & -61.00 & -61.00 & -61.00 & -61.00\\
		RBM & -7.43 & -9.56 & -5.25 & -10.17\\
		NADE & -7.19 & -10.06 & -5.48 & -10.28\\
		Note N-Gram & -10.26 & -7.91 & -4.54 & -7.50\\	
		RNN-RBM & -7.27 & -9.31 & -4.72 & -9.89\\	
		RNN\footnotemark[1] (HF) & -8.58 & -7.19 & -3.89 & -7.66\\	
		RTRBM\footnotemark[1] & -6.35 & -6.35 & -2.62 & -7.36\\	
		RNN-RBM\footnotemark[1] & -6.27 & -6.01 & -2.39 & -7.09\\	
		RNN-NADE\footnotemark[1] & -5.83 & -6.74 & -2.91 & -7.48\\	
		RNN-NADE\footnotemark[1] (HF) & -5.56 & -5.60 & -2.31 & -7.05\\
				\hline
		\bfseries{RNN-DBN} & \bfseries{-5.68} & \bfseries{-6.28} & \bfseries{-2.54} & \bfseries{-7.15}\\

		\hline
	
	\end{tabular}
	\vspace*{0.05cm}

\end{center}
\footnotetext[1]{These marked results are obtained after various preprocessing, pretraining methods and optimization techniques described in the last paragraph of this section.}

We demonstrate our technique by applying it to the task of polyphonic music generation. We used a RNN-DBN with 2 hidden DBN layers - each having 150 binary units - and 150 binary units in the RNN layer. The visible layer has 88 binary units, corresponding to the full range of the piano from A0 to C8. We implemented our technique on four datasets - \emph{JSB Chorales} , \emph{MuseData}\footnotemark[2]\footnotetext[2]{http://www.musedata.org}, \emph{Nottingham}\footnotemark[3]\footnotetext[3]{ifdo.ca/~seymour/nottingham/nottingham.html} and \emph{Piano-midi.de}.  None of the preprocessing techniques mentioned in \cite{A11} have been applied to the data, and only raw data has been given as input to the RNN-DBN. We evaluate our models qualitatively by generating sample sequences and quantitatively by using the \emph{log-likelihood} (LL) as a performance measure. Results are presented in Table 1 (a more comprehensive list can be found in \cite{A11}).

The results indicate that our technique is on par with the current state-of-the-art. We believe that the difference in performance between our technique and the current best can be attributed to lack of preprocessing. For instance, transposing the sequences in a common tonality (\emph{e.g.} C major/minor) and normalizing the tempo in beats (quarternotes) per minute as preprocessing can have the most effect on the generative quality of the model. It also helps to have as pretraining, the initialization the $W_{vh_1},W_{h_1h_2} , b_v,b_{h_1}, b_{h_2}$ parameters with independent RBMs with fully shuffled frames (\emph{i.e.} $W_{uh_1}=W_{uh_2}=W_{uv}=W_{uu}=W_{vu}=0$). Initializing the $W_{uv},W_{uu},W_{vu},b_u$ parameters of the RNN with the auxiliary cross-entropy objective via either stochastic gradient descent (SGD) or, preferably, Hessian-free (HF) optimization and subsequently finetuning significantly helps
the density estimation and prediction performance of RNNs which would otherwise perform
worse than simpler Multilayer Perceptrons \cite{A11}. Optimization techniques like gradient clipping, Nesterov momentum and the use of NADE for conditional density estimation also improve results.

\section{Conclusions and Future Work}

We have proposed a generic technique called \emph{Recurrent Neural Network-Deep Belief Network} (RNN-DBN) for modeling sequences with generative models and have demonstrated its successful application to polyphonic music generation. We used four datasets for evalutaing our technique and have obtained results on par with the current state-of-the-art. We are currently working on improving the results this paper, by exploring various pretraining and optimization techniques. We are also looking at showcasing the versatility of our technique by applying it to different problem statements. 

{\bibliographystyle{splncs}
\bibliography{egbib}
}

\end{document}